\def\year{2021}\relax
\documentclass{article} 
\pdfoutput=1
\usepackage{aaai21}  
\usepackage{times}  
\usepackage{helvet} 
\usepackage{courier}  
\usepackage[hyphens]{url}  
\usepackage{graphicx} 
\usepackage{natbib}  
\usepackage{caption} 
\title{Humans’ Assessment of Robots as Moral Regulators: Importance of Perceived Fairness and Legitimacy}
\author{Boyoung Kim and Elizabeth Phillips\\
{\normalfont George Mason University\\
Fairfax, Virginia 22030\\
\{bkim55,ephill3\}@gmu.edu}}

\begin{document}

\maketitle

\begin{abstract}
Previous research has shown that the fairness and the legitimacy of a moral decision-maker are important for people's acceptance of and compliance with the decision-maker. As technology rapidly advances, there have been increasing hopes and concerns about building artificially intelligent entities that are designed to intervene against norm violations. However, it is unclear how people would perceive artificial moral regulators that impose punishment on human wrongdoers. Grounded in theories of psychology and law, we predict that the perceived fairness of punishment imposed by a robot would increase the legitimacy of the robot functioning as a moral regulator, which would in turn, increase people’s willingness to accept and comply with the robot’s decisions. We close with a conceptual framework for building a robot moral regulator that successfully can regulate norm violations. 
\end{abstract}

\section{Introduction}
Social and moral norms are maintained and enforced through various means of moral regulation. In response to someone’s norm-violating behavior, people can verbally criticize the perpetrator or impose a penalty on the perpetrator \cite{henrich2006costly}. While this moral regulatory process has remained under the purview of humans, recent advancements in Artificial Intelligence (AI) and robotics have prompted discussions about whether AI machines, such as artificially intelligent robots, can contribute to the process of regulating norm violations. These discussions have been increasingly active in both the legal \cite{branting1998automating,sartor1998introduction,sourdin2018judge} and the everyday (non-legal) contexts \cite{briggs2015sorry,jung2015using,jackson2019tact}. To illustrate, in the legal context, there has been a controversial proposal to adopt an AI or a robot judge in court \cite{ulenaers2020impact,chen2021having,casey2020will}. A robot judge would autonomously verify the facts related to a legal case, decide the case, and determine a sentence. In the everyday context, a robot, as a member of a human-machine team, can respond to a human’s norm-violating behavior by verbally expressing its disapproval to the human transgressor \cite{briggs2015sorry,jung2015using,jackson2019tact}.

Regarding the issue about bringing artificial moral regulators into the world, some have aired concerns that human bias and discrimination may permeate through their regulatory system, which may worsen the existing inequality in societies \cite{o2016weapons,sourdin2018judge}. By contrast, others have argued that, because the artificial moral regulators would be free from biological constraints, such as hunger and fatigue, they can be less arbitrary and less biased than humans \cite{sourdin2018judge}. To build artificial moral regulators that could better the human society, it is essential that their judgments and decisions are aligned with core values humans uphold, such as the fairness of outcome \cite{von1990proportionality} and the legitimacy of a decision-maker \cite{tyler2006psychological}. When artificial moral regulators meet these conditions, people would become more likely to accept and comply with the artificial moral regulators. Resolving these ethical concerns about AI and robots functioning as moral regulators calls for interdisciplinary work between psychology, philosophy, law, and computer science.

Drawing from the literature in psychology and law, in this paper, we propose that the AI-HRI community examine people’s perception of the fairness of punishment imposed upon a human transgressor by a robot moral regulator and their perception of the legitimacy of the robot moral regulator. We also call for investigations on how the perceived fairness and the legitimacy may influence people's willingness to accept and comply with a robot’s decision to sanction human transgressors. We close with a conceptual framework of how an autonomous robot could update its moral regulatory system to distribute fair punishment as it navigates through different groups of humans.

\subsection{Legitimacy of a Moral Regulator}
Legitimacy can be defined as “a generalized perception or assumption that the actions of an entity are desirable, proper, or appropriate within some socially constructed system of norms, values, beliefs, and definitions" \cite[p. 574]{suchman1995managing}. Would people be willing to grant a similar degree of legitimacy to robots as they do to humans? It is unlikely, considering the existing findings about people’s bias against an advanced algorithm compared to a human judge in making decisions in the courtroom \cite{chen2021having}. Thus, a robot would need to earn its legitimacy as a moral regulator by demonstrating its capacities to make fair decisions.

Legitimacy can be shaped by various factors, including the perceived fairness of outcomes and procedures that lead to the outcomes \cite{tyler2021people}. One possible way for a robot to gradually acquire its legitimacy in regulating norm violations would be to show that it is capable of imposing punishment that is viewed as fair. In the next section, therefore, we discuss how the fairness of a robot moral regulator could be improved.

\subsection{Fairness of Punishment}
People’s acceptance of and compliance with an AI's decisions to sanction human transgressors would depend on whether people viewed the AI's decisions as fair \cite{volokh2018chief,chen2021having}. However, there has been mixed findings about whether people would view decisions made by AI machines as equally fair as or even fairer than decisions made by humans. For instance, Araujo et al. \shortcite{araujo2020ai} found that decisions related to justice were perceived as fairer when the decisions were led by AIs than humans, but this effect was limited to a higher impact case (i.e., making decisions on whether a criminal lawsuit should be started) than a lower impact case (i.e., making decisions on whether a parking ticket should be given). Chen et al. \shortcite{chen2021having} showed that decisions on consumer refund cases and criminal cases (e.g., bail, imprisonment) were judged as having been derived less fairly when the decision-maker was introduced as an algorithm, rather than a human judge. They also found, adding a hearing or increasing the transparency of how the decisions were reached could reduce but not completely eliminate the gap in the perceived fairness between an AI judge and a human judge.

In the present work, we explore an alternative approach to improve the perceived fairness of a robot moral regulator. Grounded in a theory of retributive justice \cite{darley2003psychology,sep-justice-retributive}, we consider that following the principle of proportionality could enhance the fairness of a robot moral regulator's decisions. The theory of retributive justice posits that, when someone commits a norm violation, they deserve punishment in return, and the intensity of the punishment should be in proportion to the severity of their violation \cite{darley2003psychology,sep-justice-retributive}. Punishment that follows the principle of proportionality tends to be viewed as fair by people \cite{miller1981social,hogan1981retributive,ball1994just}. We can thus, better understand how people assess the fairness of a robot’s punishment by studying their perception of a robot’s (vs. a human’s) punishment that either conforms or does not conform to the principle of proportionality.

Applying the principle of proportionality in distributing punishment, fair punishment would be achieved when the intensity of punishment matches the severity of a norm violation. Unfair punishment would take place in two different forms: under-punishment and over-punishment. It would be unfair to impose upon the transgressor either too weak (i.e., under-punishment) or too strong punishment (i.e., over-punishment), compared to the severity of their norm violation. A Human-Human Interaction (HHI) study \cite{wagstaff1997overpunishment} showed that, when forced to choose between under- and over-punishment that equally deviated from the fitting punishment, people's endorsement of over- and under-punishment was similar for mild violations but their endorsement of over-punishment was stronger for severe violations. Thus, the impact of over- and under-punishment on people's perception of fairness may also be influenced by various factors like the severity of a norm violation.

In a Human-Robot Interaction (HRI) study \cite{jung2015using}, it was demonstrated that, after a human teammate made an offensive comment to another human teammate, a robot could prevent the conflict from being aggravated by conveying a verbal rebuke to the offender. However, as the match between the severity of a norm violation and the intensity of the robot's rebuke was not the focus of the previous work, the question about the perceived fairness of a robot's response remains unanswered. In another HRI study \cite{jackson2019tact}, it was found that, when evaluating the harshness of a robot’s verbal response to a human’s norm-violating request, participants judged the robot’s response that was potentially more threatening to the human transgressor’s public self-image compared to the severity of the transgression as harsh. These findings suggest that people can be sensitive to the relative intensity of a robot’s verbal confrontation compared to the severity of a human’s norm violation. However, it requires further research to understand how people would evaluate the fairness of punishment delivered by a robot that addressed a norm violation caused by a human perpetrator against another human victim. Building on the previous research, we should investigate how the perceived fairness of punishment decided by a robot (as opposed to a human) influences the perceived legitimacy of the robot, and how the perceived fairness and legitimacy eventually affect people’s willingness to accept and comply with a robot moral regulator. In the next section, we introduce our working hypotheses grounded in the literature reviews we have summarized so far. 

\section{Working Hypotheses}
Based on the previous findings in human-human interactions \cite{miller1981social,hogan1981retributive,ball1994just}, we first offer our prediction for how participants’ perceived fairness of punishment would be different for the fitting punishment and the disproportionate punishment (over- and under-blaming combined).
\begin{itemize}
\item We hypothesize that, when the intensity of punishment a robot imposes on a human transgressor is proportionate to the severity of a norm violation, participants would judge the robot’s punishment as fairer, compared to when the intensity of a robot’s punishment is disproportionate.
\end{itemize}

Next, based upon the findings from human-human interactions \cite{wagstaff1997overpunishment}, we present our hypotheses about the perceived fairness of over-punishment and under-punishment as a function of the severity of a norm violation.
\begin{itemize}
\item We hypothesize that, for a severe violation, participants would judge a robot’s assigning over-punishment as fairer than assigning under-punishment.
\item We hypothesize that, for a mild violation, participants’ perception of the fairness of a robot’s punishment would not be significantly different for over-punishment and under-punishment.
\end{itemize}

Finally, we explain our prediction for the effects of the perceived fairness of punishment and the legitimacy of a robot moral regulator on participants’ willingness to comply with the robot.
\begin{itemize}
\item With repeated exposure to a robot imposing punishment that is proportional to the severity of a norm violation, participants would accumulate evidence for the robot’s capacity to decide fair punishment. These changes in the perceived fairness would increase the likelihood that participants view a robot as a legitimate moral regulator and increase their willingness to comply with a robot moral regulator in the future.
\end{itemize}

\section{A Conceptual Framework for Building a Robot Moral Regulator}
Lastly, in this section, we introduce a preliminary framework for building a robot moral regulator that may distribute fair punishment following the principle of proportionality \cite{darley2003psychology,sep-justice-retributive}.

\begin{figure}[h]
\centering
\includegraphics[width=0.45\textwidth]{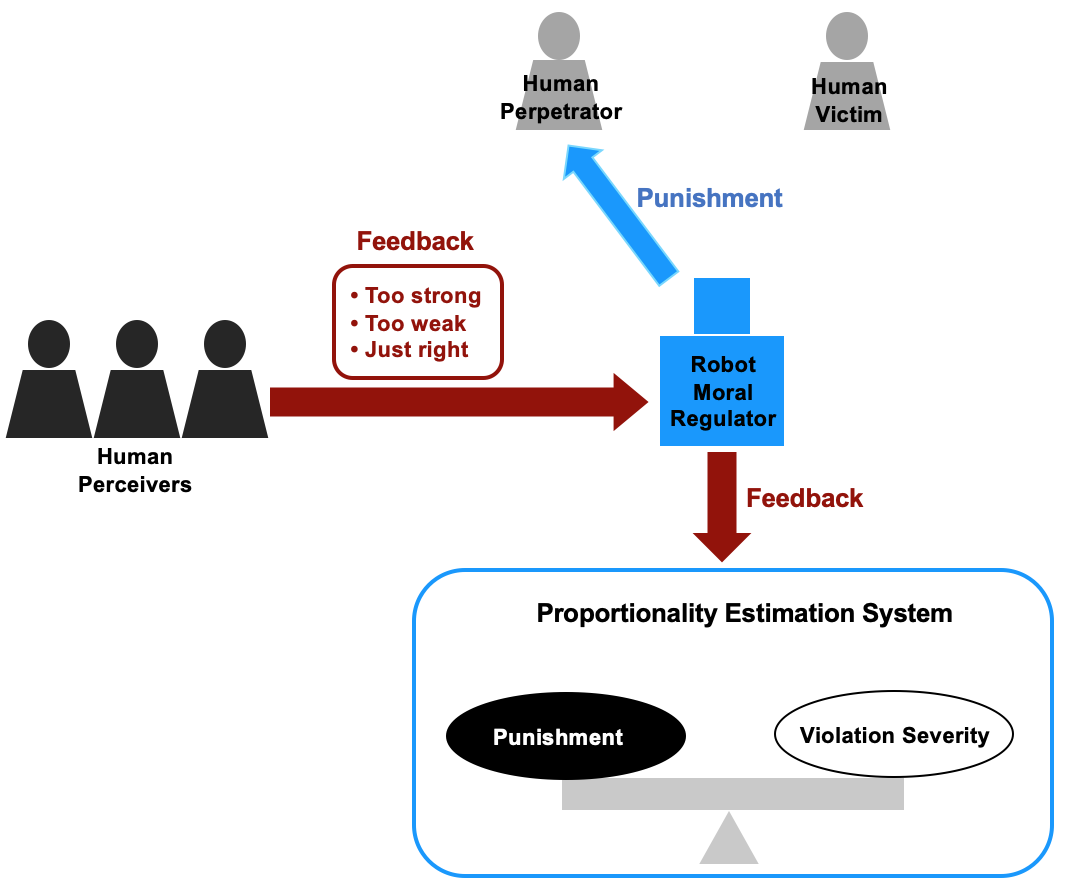}
\caption{A schematic framework of how a robot moral regulator could update its system for generating fair punishment.}
\label{fig:AIHRIFigure1}
\end{figure}

As shown in Figure~\ref{fig:AIHRIFigure1}, a robot moral regulator can be programmed to assign fair punishment that matches the severity of a norm violation caused by a human perpetrator against a human victim. Once the robot imposes punishment, it could gather feedback from third-party human perceivers on whether the punishment it imposed on the perpetrator was just right, too strong, or too weak compared to the severity of the violation. Then, the robot can update the proportionality estimation system based on the feedback. This feedback loop is critical due to the dynamic nature of norms. For instance, the norm of cooperation can dynamically change over time in different groups \cite{titlestad2019dynamic}. This implies that, when someone violates the norm of cooperation, punishment that is viewed as fitting to one group of people may not be viewed as fitting to another group. Depending on which group the transgressor belongs to, a proper and fair punishment would be different. Therefore, for a robot to be able to function as a legitimate moral regulator that successfully regulates norm violations, the robot would need to be able to flexibly adjust its proportionality estimation system.

\section{Limitations}
There are several limitations that are overlooked in this paper. First, the proposed framework does not include a monitoring system that could prevent the existing human biases and errors \cite{o2016weapons,sourdin2018judge} from being merely transferred to a robot moral regulator's decisions. The proportionality estimation system of the robot would be updated via third-party human perceivers' feedback, which may reduce the risk of decisions strictly reflecting either the victim's or the perpetrator's perspectives. However, it does not guarantee that these third-party human perceivers would be free of any biases. Second, as our discussion was focused on specific situations where victims and perpetrators of the norm-violating events are clearly determined, the proposed framework cannot explain whether and how a robot moral regulator could deal with other situations that lack such clarity. 

\section{Conclusion}
As AI systems and autonomous robots become more sophisticated, there would be more discussions about whether and how these artificially intelligent machines can be properly involved in resolving conflicts between humans. Thus, it would be essential to understand potential factors that may either increase or decrease people’s willingness to embrace a robot as a moral agent that can regulate norm violations in societies. In the current paper, we suggested that the AI-HRI research community investigate the fairness and the legitimacy as the potential factors to consider in developing well-accepted artificial moral decision-makers and proposed a conceptual framework for grounding such work. Implementations of the proposed conceptual framework into autonomous robot systems would rely upon collective efforts of the experts in various disciplines of science, including Psychology, Computer Science, and Engineering.

\section{Acknowledgment}
This work was supported in part by NSF grant IIS-1909847. We thank Tom Williams at the Colorado School of Mines for his thoughtful comments on this work.

\bibstyle{aaai21}
\bibliography{reference}

\end{document}